# The Application of Mamdani Fuzzy Model for Auto Zoom Function of a Digital Camera

*I. Elamvazuthi, P. Vasant
Universiti Technologi PETRONAS
Tronoh, Malaysia

J.Webb
University of Technology Swinnburne, Sarawak Campus,
Kuching, Sarawak, Malaysia

*Abstract*—**Mamdani Fuzzy Model is an important technique in Computational Intelligence (CI) study. This paper presents an implementation of a supervised learning method based on membership function training in the context of Mamdani fuzzy models. Specifically, auto zoom function of a digital camera is modelled using Mamdani technique. The performance of control method is verified through a series of simulation and numerical results are provided as illustrations.**

## I. INTRODUCTION

The adoption and adaptation of digital signal processing technology in consumer digital cameras has improved the performance of automatic adjustments including auto facus (AF), auto exposure (AE) and zoom tracking, which are important features for high image quality, under various shot conditions [1]. In a digital camera, optical parameters such as zoom, focus and aperture, are usually motor (computer) controlled [2].

The zoom tracking is the continuous adjustment of a camera's focal length, to keep in-focus state of camera image during zoom operation. Without zoom tracking operation, a camera system cannot maintain in-focus state during zooming. Some conventional zoom tracking techniques were implemented using curve trace stored in a look-up table [3]-[6]. The simple look-up table method, however, requires a large system memory. Moreover, it is difficult to select a proper curve when the zoom lens moves toward the tele-angle.

There are several methods that determine the variations of the intrinsic parameters developed in the literature. Recently, some research has been conducted by the employment of neural networks [7]. Nevertheless, this method is too complex to be used and require a lot of training data which might not be always available. The simplest modeling technique is to measure the intrinsic parameters using a standard offline calibration technique at several zoom and focus settings and then store the data in a look-up table [8]. This leads unfortunately to a large memory requirement especially when a precise model is needed. A novel technique using moving least-squares approach to model the variation of the camera internal parameters as a function of focus and zoom was proposed by [9]. Compared to a previous technique using a global least-squares regression scheme with bivariate polynomial functions, the new method resulted in reduction of the mean estimation error.

System Modelling based on conventional mathematical tools (e.g., differential equations) is not well suited for dealing with ill-defined and uncertain systems. By contrast, a fuzzy system employing fuzzy if-then rules can model the qualitative aspects of human knowledge and reasoning processes without employing precise quantitative analysis [10]. Increasingly, fuzzy system as a promising Computational Intelligence technique has found many industrial applications. Different fuzzy models have been developed, and successfully applied such as Mamdani model and Sugeno model. There is not a universally best fuzzy model, but each model or may have its suitable types of application. So how to automatically choose an appropriate fuzzy model or for a specific application is always important and sometimes difficult [11].

Interest has been shown recently in the applications in the fields of image recognition, especially for image zooming. Image zooming involves the adjustment of a camera's focal length to keep in-focus state of camera image during zooming operation. It is well known that there are two types of zooming operations, i.e., optical and digital zoom. Many compact digital cameras can perform both an optical and a digital zoom. A camera performs an optical zoom by moving the zoom lens so that it increases the magnification of light before it even reaches the digital sensor. In contrast, a digital zoom degrades quality by simply interpolating the image- after it has been acquired at the sensor [12]-[14].

Digital Camera involves the zooming function which allows us to focus our target image in variety of distance. Generally, the determination of zoom in (+nX) and zoom out (-nX) is proportional to the distance between the digital camera and the object that want to capture. In this paper, the auto image zooming that was implemented by using Mamdani fuzzy model is described where rules are derived from multiple knowledge sources such as previously published databases and models, existing literature, intuition and solicitation of expert opinion to verify the gathered information. This paper is not concerned with theoretical discussion of digital camera zooming functions, rather, it concentrates on the application Mamdani fuzzy model in the zooming functions of digital camera. Preliminary results of this study were published in CIM conference proceedings [15].

The rest of paper is organized as follows. A brief introduction to the concepts related to zooming functions of a digital camera is presented in Section II. Section III describes the theoretical aspects of fuzzy system by discussing the philosophy and development of fuzzy logic, fuzzy model and proposed algorithm. Section IV provides the results and discussion and lastly, conclusions are given in section V.

## II. Zoom Tracking Algorithm

In some conventional look-up table zoom tracking methods, the control circuit comprises of a microprocessor equipped with read only memory (ROM) that stores data curves for various focal distances. When the zoom lens is shifted for executing a zooming operation, the focusing lens is correspondingly shifted along the proper trace curve. Since the ROM cannot store the trace data for many distances due to limitations of its memory size, the traces between the stored traces are estimated using the following equation [16]:

$$d_c = D_c \frac{d_s}{D_s} \qquad (1)$$

where $D_s$ is the difference between the focus positions of the upper and lower traces, and $d_s$ is difference between the focus positions of the estimated and the lower trace at the zoom start point of Fig. 1. $D_c$ is the difference between the focus positions of the upper and the lower traces, and $d_c$ is the difference between the focus positions of the estimated and the lower trace at the current zoom point. As the zoom lens is shifted, the focusing lens tracks the trace stored or estimated.

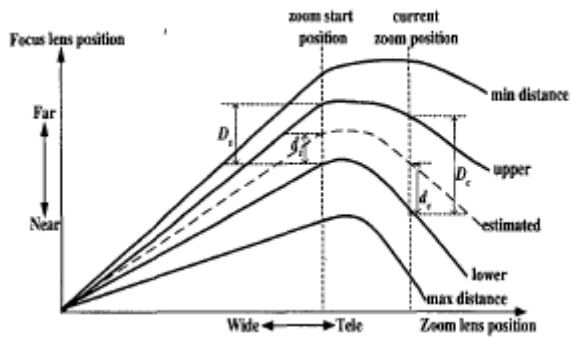

Figure 1. Conventional zoom tracking algorithm [15]

However, if the lens position sensors do not have sufficient resolution then accurate estimation curve traces are not acquired. This results in bad focusing. The de-focusing gradually increases as the zoom lens moves toward the tele-end. For better estimation more acquired data is needed thus increasing ROM size.

## III. Fuzzy Logic

### A. Philosophy and Development of Fuzzy Logic

Human reasoning is fuzzy, or approximate, and so is the real world. Fuzzy logic is the logic underlying modes of reasoning which are approximate rather than exact, thus it is closer to human reasoning and the real world than formal logic. Fuzzy logic was introduced by Zadeh in [17] and used by Mamdani to control the dynamic system in [18]. Since then, fuzzy logic has been successfully applied to many applications for automatic control (especially for non-linear ill-defined systems) [19-21]. The concept of fuzzy set is a class with unsharp boundaries. It provides a basis for a qualitative approach to the analysis of complex systems in which linguistic rather than numerical variables are employed to describe system behaviour and performance. Thus, in this work, fuzzy controller to incorporation of fuzzy techniques into the auto image zooming to achieve the automatic control during zooming function is proposed.

### B. Model Description

Fuzzy logic algorithms have been widely used in many control applications. Unlike a conventional proportional-integral-derivative (PID) controller, the FLC can achieve the goals of steady output and satisfactory transient performance simultaneously. However, choices of rule sets and membership functions significantly affect achieving these performance goals [22]. Two well-known Fuzzy rule-based Inference System are Mamdani fuzzy method and Tagaki-Sugeno (T-S) fuzzy method [23]. Advantages of the Mamdani fuzzy inference system are it's intuitive, has widespread acceptance and well suited to human cognition [24-26]. The T-S fuzzy inference system works well with linear techniques and guarantees continuity of the output surface [27-28]. But the T-S fuzzy inference system has difficulties in dealing with the multi-parameter synthetic evaluation; it has difficulties in assigning weight to each input and fuzzy rules. Mamdani model can show its legibility and understandability to the laypeople. The Mamdani fuzzy inference system shows its advantage in output expression and is used in this project.

Fuzzy logic starts with the concept of a fuzzy set. A fuzzy set is a set without a crisp, clearly defined boundary. It can contain elements with only a partial degree of membership [28]. A fuzzy set is defined by the expression below:

$$D = \{(x, \mu_D(x)) I x \in X, \mu_D(x) \in [0,1]\} \qquad (2)$$

where $X$ represents the universal set, $x$ is an element of $X$, $D$ is a fuzzy subset in X and $\mu_D(x)$ is the membership function of fuzzy set D.

Degree of membership for any set ranges from 0 to 1. A value of 1.0 represents a 100% membership while a value of 0 means 0% membership. If there are 5 subgroups of size, then 5 membership functions are required to express the size values in fuzzy rules. A membership function is a curve that defines how each point in the input space is mapped to a membership value (or degree of membership) between 0 and 1. The input space is sometimes referred to as the universe of discourse [29]. The membership functions are usually defined for input and output variables. There are some forms of membership functions such as triangular, trapezoidal, Gaussian distribution and others. In this study, triangular and trapezoidal

245    http://sites.google.com/site/ijcsis/
ISSN 1947-5500

membership functions were selected for input and output variables.

The triangular membership function is a function of a vector, x, and depends on three scalar parameters, a, b, and c, as given by an example in Fig. 2:

$$f(x;a,bc) = \mu_{Risk}^{Norm}(x = \begin{cases} 0 & x \leq a \\ \frac{x-a}{b-a} & a \leq x \leq b \\ \frac{c-x}{c-b} & b \leq x \leq c \\ \frac{-x+0.8}{0.2} & c \leq x \end{cases} \quad (3)$$

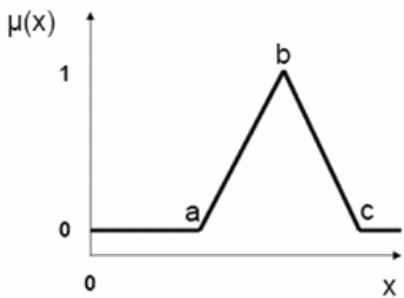

Figure 2. Schematic of triangular membership function

Fuzzy sets and fuzzy operators are the subjects and verbs of fuzzy logic. These if-then rule statements are used to formulate the conditional statements that comprise fuzzy logic. A single fuzzy if-then rule assumes the form.

IF $x$ is A THEN $y$ is B  (4)

where, A and B are linguistic values defined by fuzzy sets on the ranges (universe of discourse) X and Y, respectively. The if-then part of the rule "x is A" is called the antecedent or premise, while the then-part of the rule "y is B" is called the consequent or conclusion [28].

During data processing, fuzzy system fuzzifies the crisp data, applies the mamdani inference system using the fuzzy rules and finally, determines the zooming function of the camera through defuzzification. Defuzzification has some methods such as center of gravity (centroid (COG)), mean of maximum (MOM), and first of maximum (FOM), and so on. Center of gravity is the most popular and the most precision method for defuzzification that was used. Center of gravity method is a grade weighted by the areas under the aggregated output functions. The formula for Centroid is given as.

$$z^* = \frac{\int \mu_c(z) z \, dz}{\int \mu_c(z) \, dz} \quad (5)$$

where, $\int \mu_c(z) dz \neq 0$ for all $\mu_i$.

The formula for mean of maximum (MOM) is given as.

$$z^* = \frac{\int \mu_c(z) z \, dz}{\int \mu_c(z) \, dz} \quad (6)$$

for membership function defined as

$$\mu_c(z) = \begin{cases} 1 & \mu_c(z) = c\max \\ 0 & otherwise \end{cases}.$$

*C. Proposed Algorithm*

The objective of the proposed algorithm is to zooming function of a digital camera based on the input given as distance. Shown in Fig. 3 is a block diagram of the proposed algorithm, which includes a fuzzification block, a knowledgebase, a fuzzy inference engine and a defuzzification block.

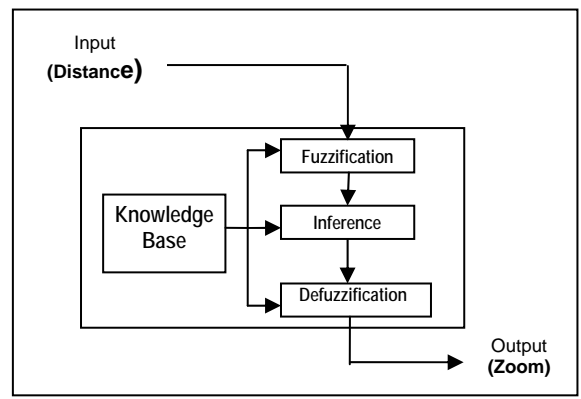

Figure 3. Fuzzy system for auto image zooming

From Fig. 3, it can be seen that the input is distance and output is zoom. The proposed algorithm consists of three steps: fuzzification, fuzzy rules, and defuzzification.

Step 1: Fuzzification.

In order to keep the number of fuzzy rules at a reasonable level, input can be defined as fuzzy set Distance with five membership functions known as Too Near, Near, Medium, Far, Too Far. For the output, Zoom fuzzy set with five membership functions such as Max Zoom Out, Min Zoom Out, Default, Min Zoom In, and Max Zoom In were defined.

Step 2: Fuzzy Rules

In this step, we use the linguistic quantification to specify a set of rules that captures the expert's knowledge about how to control the output. For example: If Distance is Too Near then Zoom is Maximize Zoom Out.

Step 3: Defuzzification

Any suitable defuzzification method can be used to defuzzify the output variable.

The inputs for fuzzy system that introduced in this study were distance that is crisp. Initially, fuzzy system fuzzifies the crisp data and then with mamdani inference system applies the fuzzy rules; finally, determines the zooming function.

The DISTANCE fuzzy set and ZOOM fuzzy set is given in Table I and II respectively.

TABLE I DISTANCE FUZZY SET

| Distance (cm) | Too Near | Near | Medium | Far | Too Far |
|---|---|---|---|---|---|
| 0 | Y* | N | N | N | N |
| 4 | Y | N | N | N | N |
| 5 | N | Y | N | N | N |
| 10 | N | Y* | N | N | N |
| 12 | N | N | Y* | N | N |
| 15 | N | N | Y | N | N |
| 20 | N | N | Y | N | N |
| 25 | N | N | N | Y | N |
| 30 | N | N | N | Y* | N |
| 35 | N | N | N | Y | N |
| 40 | N | N | N | Y | N |
| 45 | N | N | N | N | Y |
| 50 | N | N | N | N | Y* |

TABLE II ZOOM FUZZY SET

| Zoom (nX) | Max Zoom Out | Min Zoom Out | Default | Min Zoom In | Max Zoom In |
|---|---|---|---|---|---|
| -10 | Y | N | N | N | N |
| -8 | Y* | N | N | N | N |
| -6 | N | Y | N | N | N |
| -4 | N | Y | N | N | N |
| -2 | N | Y* | N | N | N |
| 0 | N | N | Y* | N | N |
| 2 | N | N | Y | Y* | N |
| 4 | N | N | N | N | N |
| 6 | N | N | N | N | N |
| 8 | N | N | N | N | Y* |
| 10 | N | N | N | N | Y |

Example rules that define the five different zooming function based on five different distance input are given in Table III.

TABLE III RULES FOR ZOOMING FUNCTION

| Rule | | Input | | Output |
|---|---|---|---|---|
| R1 | If | Distance is Too Near | then | Zoom is Maximize Zoom Out |
| R2 | If | Distance is Near | then | Zoom is Minimize Zoom Out |
| R3 | If | Distance is Medium | then | Zoom is Default |
| R4 | If | If Distance is Far | then | Zoom is Minimize Zoom In |
| R5 | If | If Distance is Too Far | then | Zoom is Maximize Zoom In |

## IV. RESULTS AND DISCUSSION

To establish the grade membership, the 'distance' is used as input and the 'zoom' is used as output. Fig. 4 and 5 illustrates the membership function of the input and output variables.

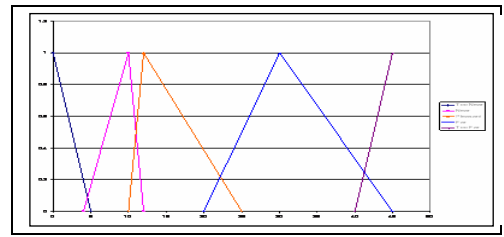

Figure 4. Membership functions of Distance

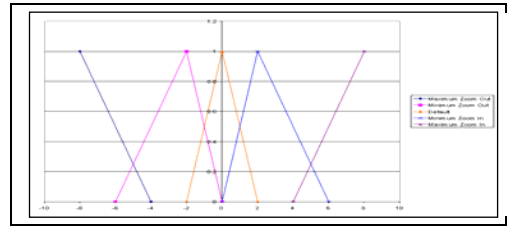

Figure 5. Membership functions of Zoom

The fuzzification of the crisp distance gives the following membership for the Distance Fuzzy set as shown in Table IV.

TABLE IV FUZZIFICATION OF DISTANCE

| | µToo Near | µNear | µMedium | µFar | µToo Far |
|---|---|---|---|---|---|
| Distance = 22 cm | 0 | 0 | 0.24 | 0.2 | 0 |
| Fire Rule | no | no | yes | yes | no |

The Inference yielded DEFAULT x MEDIUM and MINIMUM ZOOM IN x FAR. During composition, The MEDIUM and FAR sets have an output of 0.24 and 0.2 respectively as shown in Table IV.

By applying R4: If Distance is Far then Zoom is Minimize Zoom In and R3: If Distance is Medium then Zoom is Default, the output of fuzzy system can be seen in Fig. 6.

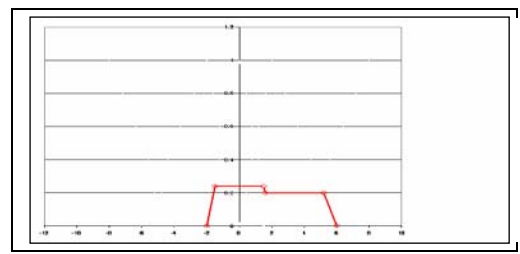

Figure 6. Composition

Defuzzification is the result of the fuzzy logic. Its primary goal is to convert the fuzzy value (an average) to a single number, a crisp value. There are many ways can be used to achieve defuzzifation and in this work, the 'Center of Area' method yielded an answer of 1.7X. The data obtained from The data from Fig. 6 is diksplayed in Table V and Table VI for defuzzification using COA and MOM methods respectively.

TABLE V  DATA AFTER COMPOSITION FOR COA

| Zoom | Default | Minimum Zoom In | Output of 2 Rules | Weighted Percentage |
|---|---|---|---|---|
| -2 | 0 | 0 | 0 | 0 |
| -1.5 | 0.24 | 0 | 0.24 | -0.45 |
| -1 | 0.24 | 0 | 0.24 | -0.3 |
| 0 | 0.24 | 0 | 0.24 | 0 |
| 1 | 0.24 | 0.2 | 0.24 | 0.3 |
| 1.5 | 0.24 | 0.2 | 0.24 | 0.45 |
| 1.6 | 0.24 | 0.2 | 0.24 | 0.32 |
| 2 | 0 | 0.2 | 0.2 | 0.4 |
| 3 | 0 | 0.2 | 0.2 | 0.6 |
| 4 | 0 | 0.2 | 0.2 | 0.8 |
| 5 | 0 | 0.2 | 0.2 | 1 |
| 5.2 | 0 | 0.2 | 0.2 | 1.04 |
| 6 | 0 | 0 | 0 | 0 |
| SUM |  |  | 2.4 | 4.16 |

TABLE VI  DATA AFTER COMPOSITION FOR MOM

| Zoom | Default | Minimum Zoom In | Output of 2 Rules | Weighted Percentage |
|---|---|---|---|---|
| -1.5 | 0.24 | 0 | 0.24 | -0.45 |
| -1 | 0.24 | 0 | 0.24 | -0.3 |
| 0 | 0.24 | 0 | 0.24 | 0 |
| 1 | 0.24 | 0.2 | 0.24 | 0.3 |
| 1.5 | 0.24 | 0.2 | 0.24 | 0.45 |
| 1.6 | 0.24 | 0.2 | 0.24 | 0.32 |
| SUM |  |  | 1.44 | 0.32 |

Defuzzification using COA method shows that the computation leads to a SINGLE VALUE for the size, which is an average value computed with respect to the centre of gravity of the output fuzzy set: 4.16/2.44 = 1.7. Defuzzification using MOM method shows that the computation leads to a SINGLE VALUE for the size, which is the output fuzzy set: 0.32/1.44 = 0.22. The difference between these two methods is too big. Simulation results show that this is true for different set of values. It can be deduced that the use of centre of area method produces more accurate results compared to the mean of maxima method.

## V.  CONCLUSION

In the study, the auto image zooming function has been enhanced using Mamdani fuzzy model. We restricted our study to the case of controlling the zoom function using distance as input. The performance of our algorithm is quite good because it is capable of discriminating the zoom functions based on distance. However, our simulations show that some improvements and extensions can be made to our approach. In future, advanced parameter initialization methods and learning algorithms such as genetic algorithm, particle swarm optimization can be applied in improving the performance of zooming function.

ACKNOWLEDGMENT

The authors would like top thank Universiti Teknologi PETRONAS and University of Technology Swinburne, Sarawak Campus, Kuching, Sarawak, Malaysia for supporting this work.

AUTHORS PROFILE

I. Elamvazuthi is a lecturer at the Department of Electrical and Electronic Engineering of Universiti Teknologi PETRONAS (UTP), Malaysia. His research interests include Control Systems, Mechatronics and Robotics.

P. Vasant is a lecturer at the Department of Fundamental and Applied Sciences of Universiti Teknologi PETRONAS (UTP), Malaysia. His research interests are Soft Computing and Computational Intelligence.

J. Webb is a lecturer at the University of Technology Swinburne, Sarawak Campus, Kuching, Sarawak, Malaysia. He specializes in Computational Methods, Nano-Physics and Mechatronics.